\documentclass{article}
\usepackage{arxiv}

\usepackage[utf8]{inputenc} 
\usepackage[T1]{fontenc}    

\usepackage{array, textcomp, url, comment, tabularx} 
\usepackage{graphicx, pdflscape}

\usepackage[table,xcdraw,dvipsnames]{xcolor}

\usepackage{amsmath, amssymb, amsthm, amsfonts, mathtools, stmaryrd, pifont, bbm} 
\usepackage{algorithm, algorithmicx}
\usepackage[noEnd=true, 
            commentColor=black!70, 
            italicComments=true]{algpseudocodex}
\usepackage{enumitem, booktabs, multirow, multicol}
\usepackage[compress]{cite}
\usepackage{nicefrac}       

\usepackage{etoolbox}
\AtBeginEnvironment{table}{\footnotesize}
\AtBeginEnvironment{tabular}{\footnotesize}
\AtBeginEnvironment{figure}{\footnotesize}

\usepackage[caption=false,listofformat=empty]{subfig}

\newcommand{\withOrcid}[2]{%
    \begingroup%
    \hypersetup{urlcolor=black}%
    \ifstrempty{#2}{#1}{\href{https://orcid.org/#2}{%
    {#1}\includegraphics[scale=0.06]{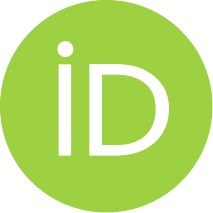}}}%
    \endgroup%
}

\usepackage{hyperref}
\hypersetup{
 colorlinks=true,
 linkcolor=blue,
 citecolor=blue,
 filecolor=magenta, 
 urlcolor=blue,
 pdfauthor = {Filus, K. and Cruz-Duarte, J. M.},
}

\begin{document}
\frenchspacing

\title{
    Semantically Guided Adversarial Testing of Vision Models Using Language Models%
    \thanks{This work was supported by the~START Scholarship of the Foundation for Polish Science (FNP) for outstanding young scholars, agreement No START 017.2025.}
}
\def\shorttitle{%
Semantically Guided Adversarial Testing of Vision Models Using Language Models
}

\author{%
    \withOrcid{%
    Katarzyna Filus}{0000-0003-1303-9230}\\
    Institute of Theoretical and Applied Informatics,\\ Polish Academy of Sciences, Gliwice, Poland\\
    \texttt{kfilus@iitis.pl}
    \And%
    \withOrcid{%
    Jorge M. Cruz-Duarte}{0000-0003-4494-7864}\\
    University of Lille, CNRS, Inria, Centrale Lille,\\ UMR 9189 CRIStAL, F-59000 Lille, France\\
    \texttt{jorge.cruz-duarte@univ-lille.fr}\\
}

\maketitle

\begin{abstract}

In targeted adversarial attacks on vision models, the selection of the target label is a critical yet often overlooked determinant of attack success. 
This target label corresponds to the class that the attacker aims to force the model to predict.
Now, existing strategies typically rely on randomness, model predictions, or static semantic resources, limiting interpretability, reproducibility, or flexibility. 
This paper then proposes a semantics-guided framework for adversarial target selection using the cross-modal knowledge transfer from pretrained language and vision-language models. 
We evaluate several state-of-the-art models (BERT, TinyLLAMA, and CLIP) as similarity sources to select the most and least semantically related labels with respect to the ground truth, forming best- and worst-case adversarial scenarios. 
Our experiments on three vision models and five attack methods reveal that these models consistently render practical adversarial targets and surpass static lexical databases, such as WordNet, particularly for distant class relationships. 
We also observe that static testing of target labels offers a preliminary assessment of the effectiveness of similarity sources, \textit{a priori} testing. 
Our results corroborate the suitability of pretrained models for constructing interpretable, standardized, and scalable adversarial benchmarks across architectures and datasets.

\smallskip
All codes and results used in this work can be found in the repository: \url{https://github.com/kafilus/language-vision-attacker}.

\smallskip
\textit{This manuscript is going to be submitted to a journal for possible publication. Copyright may be transferred without notice, after which this version may no longer be accessible.}

\end{abstract}

\keywords{%
    Artificial Intelligence Security \and 
    Adversarial Attacks \and 
    Target Label Selection \and 
    Neural Network Testing \and 
    Computer Vision \and 
    Semantic Similarity.
}

\section{Introduction}
\lettrine{O}{ver} the past decade, a wide range of adversarial attacks have been developed to expose vulnerabilities in deep learning models in computer vision \cite{bib:szegedy2013intriguing, bib:goodfellow2014explaining, bib:carlini2017towards, bib:byun2022improving, bib:filus2023netsat}. These vulnerabilities pose a distinct but equally important security threat alongside traditional cyberattacks \cite{bib:gelenbe2025adaptive, bib:filus2020random, bib:guerra2020medbiot} and software vulnerabilities \cite{bib:siavvas2024transforming, bib:filus2023software, bib:filus2020randomvp, bib:hanif2022vulberta}, by targeting model behavior and knowledge. The field of adversarial attacks has significantly advanced in crafting perturbations that effectively change models' predictions while being imperceptible. However, as attack techniques have matured, the focus should shift from solely inventing new attack types to systematically evaluating and benchmarking attacks and models. 
A critical yet often-disregarded factor is how the target labels are selected for targeted attacks. This can cause severe consequences by deceiving networks into predicting designated target classes \cite{bib:byun2022improving}. Most studies either choose random target classes or use the predicted classes' probabilities, ignoring the semantic meaning and structure behind label relationships \cite{bib:kurakin2016adversarial, bib:carlini2017towards, bib:hu2021tkml}.

\begin{figure}[!htp]
  \centering
    \includegraphics[
    width=0.8\linewidth]{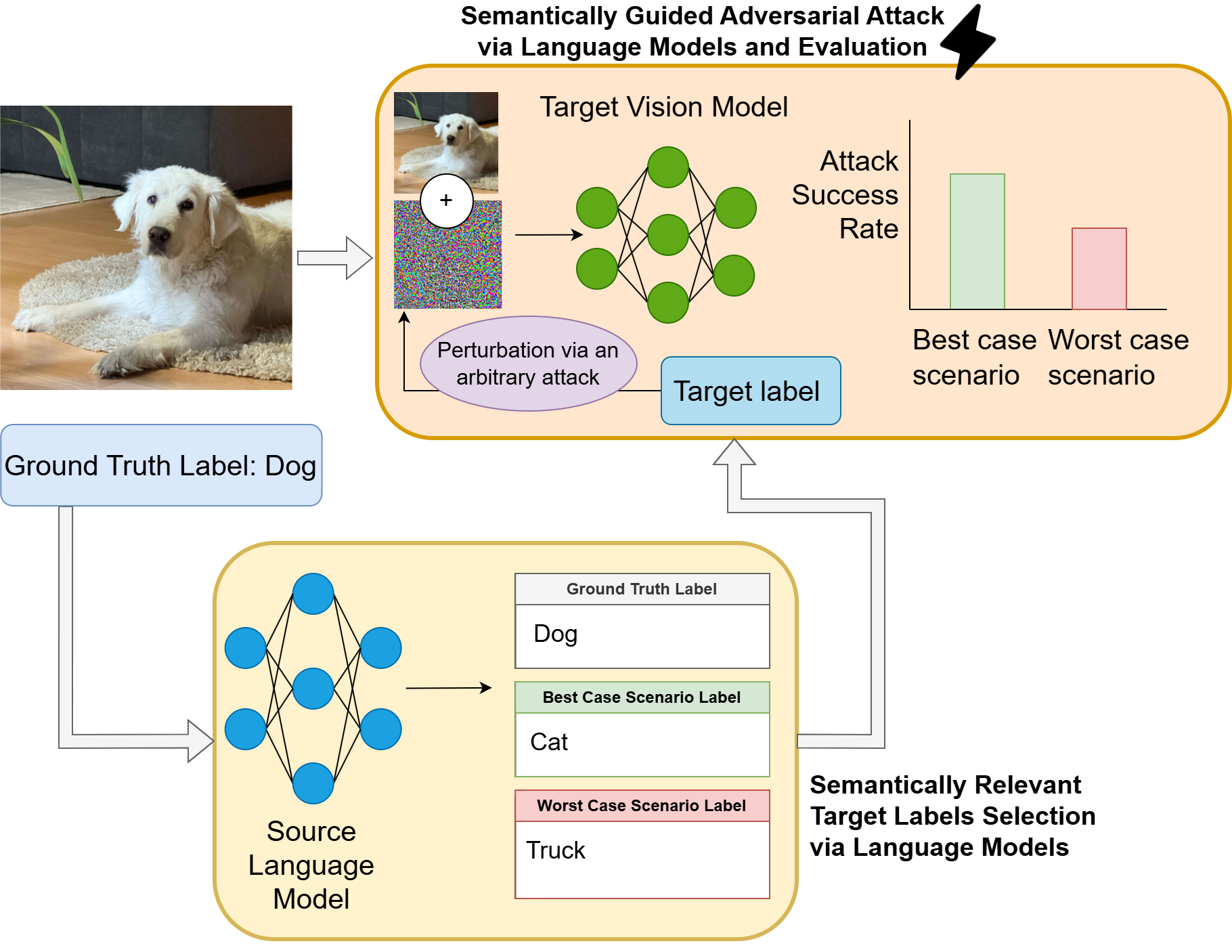}
  \caption{Overview of our target label selection method. We propose using the semantics embedded in trained language models to guide adversarial testing by selecting the best (easiest) and worst (hardest) test case scenarios.
  }
  \label{fig:overview}
\end{figure}

The current label-choice strategies are problematic to interpret and explain, which is crucial for testing transparency. For instance, when using the least-likely predicted classes, the selected target labels may depend on unpredictable, image-specific artifacts, such as co-occurring objects or spurious background patterns, or even label noise. Therefore, it is unattainable to easily determine why a particular class is considered ``least likely'' for a given input, nor whether the target is semantically relevant or irrelevant. Since this selection method is tied to the model's current predictions, it varies across individual images, making it hard to verify the validity or plausibility of each target. This lack of transparency and scalability makes image-dependent strategies unsuitable for systematic benchmarking.

To build meaningful and security-relevant adversarial benchmarks, we assert that target label selection strategies need standardization and have to be rooted in interpretable approaches offered, e.g., via semantics. Recently, researchers proposed a similarity-driven strategy for target label selection based on class representations embedded in network weights and static language databases \cite{bib:filus2024similarity}. Selecting targets based on network weights and their perceived class space is useful to understand how a particular network organizes classes and evaluate individual models. However, the internal similarity structure of different models varies, making it unsuitable as a universal reference point for cross-architecture comparison. Therefore, we aim to establish a semantic reference that is agnostic to a target model, facilitating such evaluation. While using the semantics of static lexical databases, such as WordNet \cite{bib:filus2024similarity}, is a step in the right direction, it brings practical and technical disadvantages. WordNet similarity requires manually mapping class labels to specific nodes. While reducing ambiguity, this process is time-consuming, error-prone, and challenging to scale, particularly for datasets with myriad classes. In addition, some parts of WordNet are less refined, affecting similarity metrics based on the normalization of the node height \cite{bib:filus2025semantic}. Moreover, WordNet's tree-like nature leads to discrete and often repeated similarity values, specifically among distant classes. This can cause ambiguity when ranking candidate labels by similarity, which in turn limits the precision of target selection and is dependent on the greedy nature of sorting algorithms.

We propose a unified framework for semantics-guided target label selection in adversarial attacks, leveraging language and vision-language models to compute semantic similarity between classes. Target classes are selected based on their similarity to the ground truth, with the most similar ones representing ``best-case'' targets, those that are easier to reach, and the most dissimilar ones representing ``worst-case'' targets, posing a greater challenge due to their semantic distance. We present a simple overview of this approach in \figurename{}~\ref{fig:overview}. Unlike model-dependent and image-dependent target selection strategies, our similarity-based strategy is reproducible and decoupled from instance-specific confounding factors. It is also interpretable because it is rooted in semantics. This approach allows us to create target look-up tables that are constant throughout evaluation \cite{bib:filus2024similarity}. To prove it, we compare our method with WordNet-based similarity and evaluate its effectiveness using standard adversarial attack techniques on a common attack benchmarking dataset, i.e., NIPS 2017 Adversarial Competition DEV \cite{bib:nips2017dataset}. We use three state-of-the-art text and text-image models as similarity sources (BERT \cite{bib:devlin2018bert}, TinyLLAMA \cite{bib:zhang2024tinyllama}, and CLIP \cite{bib:radford2021learningCLIP}) and conduct five attack types on three standard vision models.

Our experiments reveal that using source models from the text domain offers superior results for the worst-case scenario. They also show that text-image models can provide comparable results to WordNet regarding the best-case scenario. 
The main contributions of this work are:
\begin{enumerate}[label=\itshape\Alph*.]
    \item An evaluation of the reliability of text and vision language models for generating meaningful adversarial targets;
    \item A comparison between pretrained semantic models and WordNet in both best- and worst-case scenarios;
    \item An analysis of the impact of target similarity on the severity of attacks based on post-attack semantic deviation; and
    \item Evidence that static similarity computations between class labels anticipate the behavior of models under attack.
\end{enumerate}
As our proposal is more flexible than static language databases as a similarity source, this shift toward semantically grounded adversarial testing constitutes a crucial step in standardizing security evaluation protocols. It also enables consistent and interpretable benchmarking across images, architectures, and datasets. These are critical aspects for the security of intelligent computer systems.

\section{Related Work}

There is a plethora of works on adversarial attacks in vision primarily focusing on crafting effective perturbations \cite{bib:li2024adversarial, bib:madry2017towards, bib:carlini2017towards, bib:szegedy2013intriguing, bib:moosavi2017universal, bib:goodfellow2014explaining, bib:chen2024content}, techniques to obtain transferability between models \cite{bib:zoph2018learning, bib:zhou2018transferable, bib:wang2023improving} and attack mitigation \cite{bib:filus2024evaluating, bib:bayat2024adversarial}. In contrast, minimal attention has been paid to target label selection in targeted attacks, despite their central role in the interpretability, reproducibility, and standardization of adversarial testing. 
The most common strategies include choosing random labels \cite{bib:kurakin2016adversarial, bib:carlini2017towards, bib:hu2021tkml} and relying on probabilities of classes predicted for a given image \cite{bib:kurakin2018adversarial}. In the literature, works implement predictions with the lowest or highest probability labels that are not the ground truth, i.e., Least Likely or Most Likely target labels \cite{bib:kurakin2018adversarial, bib:hu2021tkml}. Alternatively, the $K$ least/most likely labels can be used \cite{bib:hu2021tkml}. While these approaches are simple to implement, they inherently depend on the model or instance. This makes the resulting myriad target labels unfeasible to interpret and compare across models. Moreover, the probability-based methods are sensitive to factors such as spurious background patterns or object co-occurrence, making it even more unclear why a target label was selected for a specific image. Such a lack of transparency and scalability makes image-dependent strategies unsuitable for systematic benchmarking. 

Besides, some works reveal the potential of semantics for the adversarial attacks domain. For example, Mahmood and Elhamifar proposed an optimization approach for multi-label learning that modifies the predictions of desired labels while ensuring other labels will not get affected \cite{bib:mahmood2024semantic}. Some authors have considered semantics and other similarity sources to build evaluation metrics in adversarial testing \cite{bib:filus2024similarity, bib:mopuri2020adversarial, bib:filus2023netsat}. 
While these works reveal the potential of incorporating similarity and semantics for adversarial testing, they focus on different aspects, such as metrics and multi-label optimization.
In contrast, we propose to use similarity to guide adversarial testing efficiently.

One recent work introduced a method for target label selection that exploits similarities derived from internal model representations embedded in weights and lexical databases like WordNet, to define best- and worst-case testing scenarios~\cite{bib:filus2024similarity}.
In contrast to this work, we propose to transfer the knowledge of trained text and text-image neural networks to guide testing in a cross-modal manner. We utilize the embeddings of the known classes, generated with the text-only representation of labels, to estimate similarity between concepts. We exploit these characteristics to construct the best and worst adversarial cases by choosing the most and least similar target labels. This makes our procedure more flexible and more feasible to automate. This is because using WordNet as a source of similarities can be troublesome, as it requires manual mapping of labels. Besides, there are many same-valued similarities and deficiencies in the scope of some parts of these databases. Still, utilizing target network weights makes the testing inconsistent when different target models are compared, limiting reproducibility. Therefore, this is the first such approach to adversarial target label selection. We reference the WordNet-based approach presented in  \cite{bib:filus2024similarity}.

\section{Methods}

\subsection{Semantic Similarity Reference}

Semantic similarity is a relation between terms or objects with a similar meaning. It can be measured, e.g., via static lexical databases with hierarchical, tree-like structures, such as WordNet \cite{bib:kolb2009experiments}. The assumption is that the terms to be analyzed are available as the nodes of a given structure. Thus, semantic similarity via WordNet can be measured through the path length \cite{bib:pedersen2004wordnet}, Wu and Palmer (WUP) \cite{bib:wu1994verb}, Leacock and Chodorow \cite{bib:leacock1998combining}, and so forth. In particular, WUP has been widely used to describe semantic similarity between classes \cite{bib:mopuri2020adversarial, bib:filus2024similarity}. Therefore, we implement this measure to determine our reference best- and worst-case scenarios. 

For a given ground truth class \( c_{\text{gt}} \in \mathcal{C} \), we define the \textit{Most Similar (MS)} variant or the best-case scenario class \( c^+ \) as the class with the highest WUP semantic similarity to \( c_{\text{gt}} \), and the \textit{Least Similar (LS)} variant or the worst-case target \( c^- \) as the class with the lowest WUP. These can be computed as follows,
\begin{equation}
    c^+ = \underset{c \in \mathcal{C} \setminus \{c_{\text{gt}}\}}{\operatorname{argmax}} \left\{\text{WUP}(c_{\text{gt}}, c)\right\}, 
\quad\text{and}\quad
    c^- = \underset{c \in \mathcal{C} \setminus \{c_{\text{gt}}\}}{\operatorname{argmin}} \left\{\text{WUP}(c_{\text{gt}}, c)\right\}.
\end{equation}
We precompute and store \( (c^+, c^-) \) pairs for all \( c_{\text{gt}} \in \mathcal{C} \) in a lookup table to enable consistent, interpretable, and efficient target label selection during adversarial testing.

\subsection{Semantic Similarity via Pretrained Models}

To compute semantic similarity between classes without relying on static language knowledge bases, such as WordNet, we propose using text-only embeddings computed via different language models (i.e., BERT and TinyLLAMA) and image-language models (i.e., CLIP) trained on large datasets in a self-supervised manner. We refer to these language and image-language models as similarity source models. For each class \( c \in \mathcal{C} \), we consider its textual name (e.g., ``\texttt{school bus}'') and encode it using pretrained models. Let \( \mathbf{e}(c) \in \mathbb{R}^d \) denote the resulting embedding for class \( c \). Subsequently, the semantic similarity between two classes is assessed using the cosine similarity function, \(\operatorname{cs}(\cdot)\). So, given a ground truth class \( c_{\text{gt}} \), the {MS variant} \( c^+ \) and the {LS variant} \( c^- \) are defined respectively as
\begin{equation}
    c^+ = \underset{c \in \mathcal{C} \setminus \{c_{\text{gt}}\}}{\operatorname{argmax}} \left\{\operatorname{cs}(c_{\text{gt}}, c)\right\},
\quad\text{and}\quad
    c^- = \underset{c \in \mathcal{C} \setminus \{c_{\text{gt}}\}}{\operatorname{argmin}} \left\{\operatorname{cs}(c_{\text{gt}}, c)\right\}.
\end{equation}
As with the WUP computation, we precompute and store the \( (c^+, c^-) \) pairs for all classes using each language model, creating a lookup table (one per source model). This must be done once, before the adversarial training, and therefore does not incur any significant costs during testing. Such an approach supports semantics-driven and consistent testing, thus making security evaluations more interpretable and reproducible. This process avoids manual taxonomy mapping and provides continuous similarity scores, which is the most important for more distant classes. Therefore, the procedure is well-suited for large-scale security evaluations.

\subsection{Attack Evaluation Metrics}

We use two standard metrics to evaluate the success of targeted attacks: Fooling Rate (FR) and Targeted Success Rate (TSR). These are computed for the vision models under attack, namely, the attack target models.
On the one hand, FR measures the percentage of images for which the attack caused the label change, regardless of what change it triggered, such as
\begin{equation}
    \text{FR} = \frac{1}{N}\sum^N_{i=1} \mathbbm{1}(\hat{y}_i^\text{pre} \neq \hat{y}_i^\text{post}).
\end{equation}
On the other hand, TSR measures how many targets have been reached by executing the attack and can be defined as follows,
\begin{equation}
    \text{TSR} = \frac{1}{N}\sum^N_{i=1} \mathbbm{1}(\hat{y}_i^\text{target} \neq \hat{y}_i^\text{post}).
\end{equation}
In both cases, \(\hat{y}_i^k\), \(\forall k\in\{\text{pre},\text{post},\text{target}\}\), correspond to the \(i\)\textsuperscript{th} pre-attack, post-attack, and target labels, and \(N\) is the number of samples in the tested dataset.

We also use Dissimilarity Metric (DM), which was introduced in \cite{bib:filus2023netsat}, to more thoroughly measure the impact of different label choosing strategies on the model predictions. Instead of calculating the attack success on a binary scale, it determines how distant the post-attack predicted labels are in the model's perceived space from the ground truth labels. It does that based on the similarity of classes expressed as templates within the target model's weights. DM takes values in the interval $[0, 1]$, with zero reflecting 100\% standard accuracy of the model, and one meaning that all post-attack predictions are different from the ground truth and maximally severe from the model perspective. All values in between indicate an increasing severity.  It can be treated as a mean-normalized rank distance between labels. This is helpful for a more thorough evaluation of attack success than binary TSR, as it estimates the degree of damage caused by the attack, even if the target label was not reached. This metric is paramount for the LS variant evaluation (DM should be as high as possible) and the MS variant (as low as possible, but non-zero, with approximately 0.001 being the perfect value for ImageNet).

Moreover, we also calculated the values of DM in a `static' setup. Namely, we computed the dissimilarity between the ground truth and target values (and not the predicted values, as in the standard setup) obtained for different similarity source models. Thus, the mean is computed in this setup over all ImageNet classes, not instances in the dataset. Suppose the standard and dynamic DM results are similar. In that case, it means that we can assess the potential vulnerability of the model to given targets and similarity sources with no image instances \textit{a priori} to testing.

\subsection{Experimental Setup}

We conducted experiments with five different attacks: FGSM \cite{bib:goodfellow2014explaining}, C\&W \cite{bib:carlini2017towards}, PGD \cite{bib:madry2017towards}, Momentum Iterative Method (MIM) \cite{bib:dong2018boosting} and Simultaneous Perturbation Stochastic Approximation (SPSA) \cite{bib:uesato2018adversarial}. We utilized three state-of-the-art similarity source models, two from the text domain, BERT \cite{bib:devlin2018bert} and TinyLLAMA \cite{bib:zhang2024tinyllama}; and one from the text-image domain, CLIP \cite{bib:radford2021learningCLIP}. The models are taken from the Hugging Face repository\footnote{Hugging Face repository: \url{https://huggingface.co/models}} and implemented in PyTorch. We also employed three vision models as targets for the attacks, such as MobileNetV2 \cite{bib:sandler2018mobilenetv2}, EfficientNetV2B0 \cite{bib:tan2021efficientnetv2}, and ResNet50V2 \cite{bib:he2016identity}. These models are freely available via Keras Application\footnote{Keras Application: \url{https://keras.io/api/applications/}} and were all trained on a common vision benchmark large-scale dataset, i.e., ImageNet \cite{bib:deng2009imagenet, bib:russakovsky2015imagenet}. We used the NIPS 2017 Adversarial Competition dataset \cite{bib:nips2017dataset} for tests, which is a common benchmark employed for adversarial attacks \cite{bib:byun2022improving, bib:li2020towards}, compatible with ImageNet-trained networks.
Moreover, all codes and results of this work can be found in the freely accessible repository at \url{https://github.com/kafilus/language-vision-attacker}.

A portion of the experiments was carried out using the Grid'5000\footnote{Grid'5000 website: \url{https://www.grid5000.fr}} testbed, supported by a scientific interest group hosted by Inria and including CNRS, RENATER, several Universities, and other organizations. In particular, the machine (chifflot-5) features an NVIDIA Tesla P100 GPU with 16~GB of RAM. 
It operates in a Linux-based environment, employing 13 allocated nodes as part of the Grid'5000 Lille site's infrastructure. The other experiments were performed on a Windows-based workstation with an NVIDIA TITAN RTX GPU and 64~GB of RAM.

\section{Experimental Results}

This section presents the results of our experiments designed to assess the effectiveness of semantics-guided target label selection via language models for adversarial vision testing. The analysis is structured around four key research questions: 
\begin{enumerate}[label=\textit{\Alph*}.]
    \item Can text domain models be reliably used to construct best- and worst-case adversarial testing cases?
    \item How do text and vision-language models compare to static semantic resources like WordNet in selecting target labels for attacks? 
    \item Which similarity sources impact attack severity more, as measured by dissimilarity in model predictions?
    \item Which similarity sources provide targets with the highest static compatibility with vision networks?
\end{enumerate}
The following subsections present the related results and provide detailed answers.

\subsection{Reliability of Building Adversarial Cases with Text Models} 

\figurename{}~\ref{fig:bar-plots_MSLS} presents the results of different similarity sources for the MS and LS case scenarios. For the MS case (\figurename{}~\ref{fig:bar_plots_MS}), all similarity sources obtained higher targeted and non-targeted success results than for the LS variant (\figurename{}~\ref{fig:bar_plots_LS}), which is a desirable outcome. 
The compact upper-end distributions in the MS violin plots visually support this, indicating more consistent and successful attacks.
\tablename{}~\ref{tab:fr_tsr_std_min_max} complements \figurename{}~\ref{fig:bar-plots_MSLS} with a more detailed panorama of the results based on the Average (Avg.), Standard Deviation (St. Dev.), Minimum (Min.), and Maximum (Max.) values of FR and TSR per the source model.
For the MS variant, all sources render high FR values (Avg. $\geq$ 0.7961), confirming that label changes are easy to induce when the target is semantically close. TSR values, however, vary more. WUP and CLIP achieve the highest average TSR values (0.5563 and 0.5529), followed by BERT and LLAMA (0.4770 and 0.4592), suggesting that WordNet and vision-language models define slightly more accessible best-case targets. 
Among the similarity sources, CLIP and WUP typically exhibit the lowest variability, indicating greater stability and reliability in the test scenarios they produce. 
In the LS case, FR values are distributed similarly, with average values from 0.7745 to 0.7971, obtained via CLIP and WUP, respectively.
In particular, WUP presents the least varied values, with a St. Dev of 0.3495, not so closely followed by BERT, which has 0.3759.
Notably, distributions observed in \figurename{}~\ref{fig:bar_plots_LS} get their shape from three samples below the 0.5 mark, except for LLAMA with two, and then a denser concentration on larger FR values.
Besides, TSR drops markedly across all sources, with means ranging from 0.3919 to 0.4377, corroborating the increased difficulty of hitting distant semantic targets. CLIP yields the lowest average TSR, followed very closely by LLAMA, indicating that its worst-case selections are maximally challenging.
Yet, slightly higher dispersion is observed in the LS variant, especially for CLIP, which reflects more variability in attack success when targets are semantically distant.
CLIP and LLAMA exhibit very similar distributions regarding central tendency and dispersion in both FR and TSR. This alignment suggests that, despite their different architectures and modalities, both models capture class-level semantic relationships closely.

\def\variant{_vaginas}
\def\wpr{0.35}
\begin{figure}[!htpb]
  \centering
  \subfloat[Most Similar (MS)\label{fig:bar_plots_MS}]{%
  \centering
    \includegraphics[
    trim={8pt 10 10pt 10pt}, clip, 
    width=\wpr\linewidth]{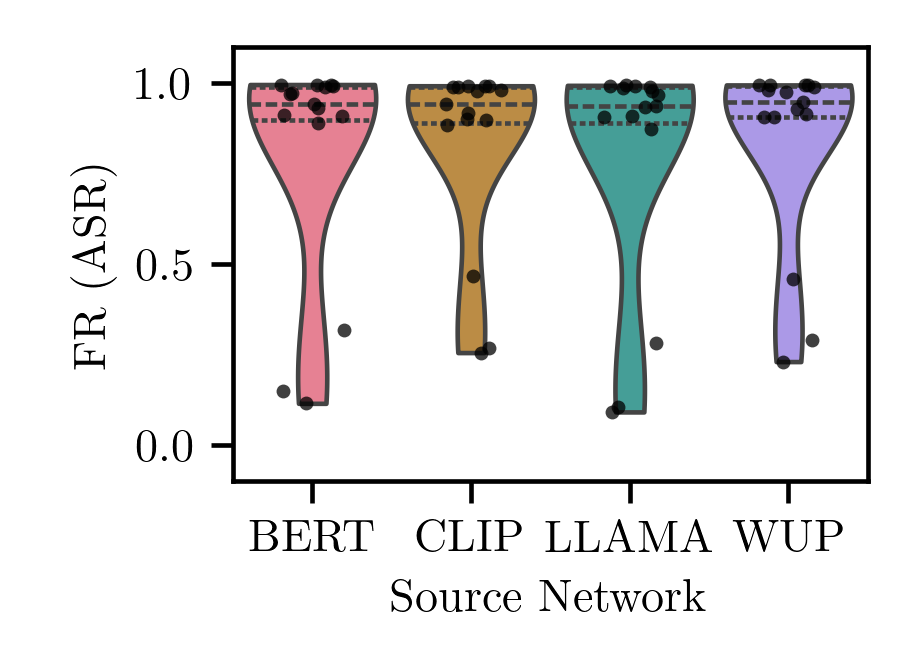}%
    \includegraphics[
    trim={8pt 10 10pt 10pt}, clip, 
    width=\wpr\linewidth]{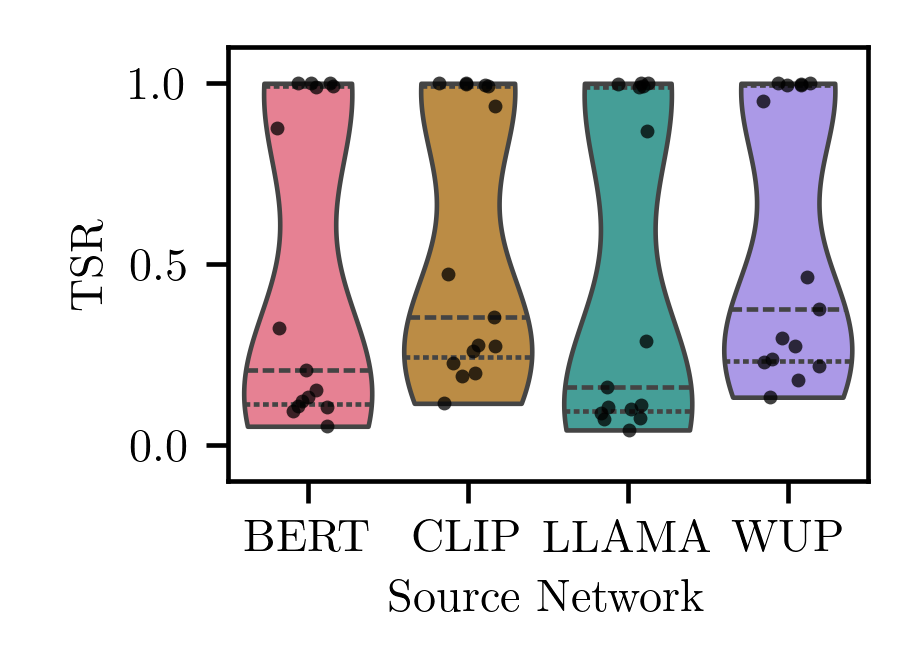}
  }\\
  \subfloat[Least Similar (LS)\label{fig:bar_plots_LS}]{%
  \centering
    \includegraphics[
    trim={8pt 10 10pt 10pt}, clip, 
    width=\wpr\linewidth]{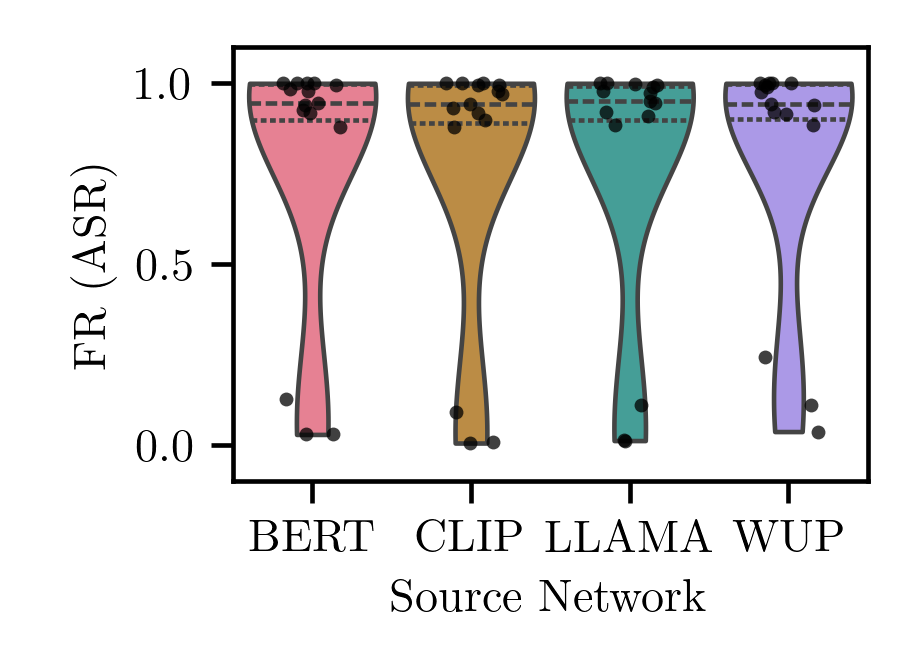}%
    \includegraphics[
    trim={8pt 10 10pt 10pt}, clip, 
    width=\wpr\linewidth]{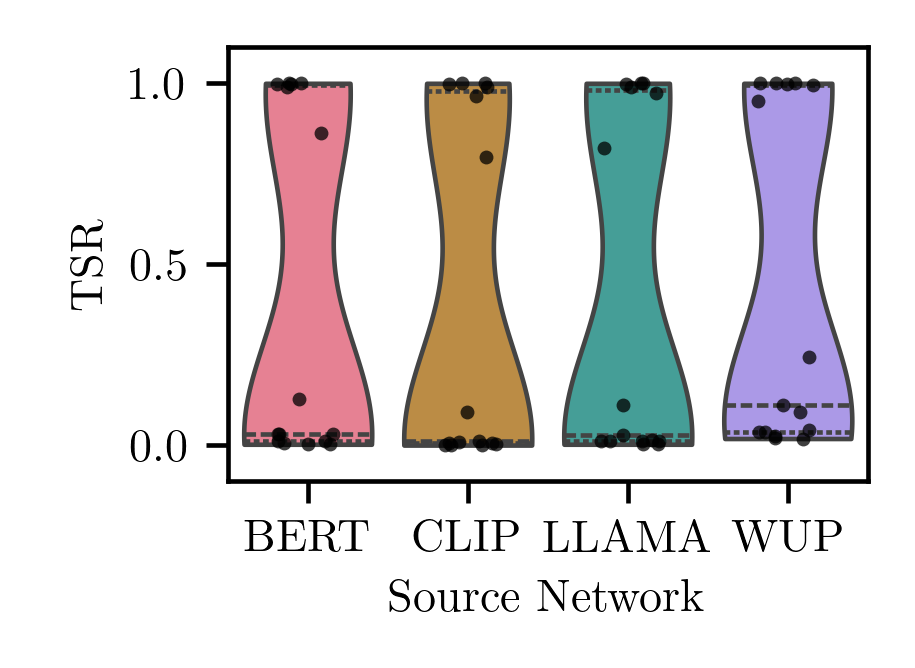}
    %
  }
  \caption{Aggregated attack performance for (a) MS and (b) LS target selection strategies across all semantic similarity sources. Fooling Rate (FR) indicates overall label change success, while Targeted Success Rate (TSR) reflects the success in reaching the designated targets. For MS cases, higher values imply easier-to-reach adversarial targets; for LS cases, lower values indicate more challenging ones. Dashed lines denote the quartile ranges, and black dots represent individual samples.
  }
  \label{fig:bar-plots_MSLS}
\end{figure}

\begin{table}[ht]
\centering
\caption{Aggregated statistics of Fooling Rate (FR) and Targeted Success Rate (TSR) for both Most Similar (MS) and Least Similar (LS) variants. For MS cases, higher values imply easier-to-reach adversarial targets, while for LS cases, lower values indicate more challenging targets. The best results per row are highlighted in bold across the sources. 
}\label{tab:fr_tsr_std_min_max}
\begin{tabular}{@{}*{4}{l}*{3}{c}@{}}
\toprule
\textbf{Metric} & \textbf{Variant} & \textbf{Statistic} & \textbf{BERT} & \textbf{CLIP} & \textbf{LLAMA} & \textbf{WUP} \\
\midrule
\multirow{7}{*}{FR} 
&\multirow{4}{*}{MS} 
    & Avg.        & 0.8052          & 0.8302            & 0.7961            & \textbf{0.8343} \\  
    && St. Dev.    & 0.3209          & \textbf{0.2653}   & 0.3342            & 0.2684 \\
    && Min.       & 0.1154          & \textbf{0.2557}   & 0.0916            & 0.2308 \\
    && Max.       & \textbf{0.9966} & 0.9932            & 0.9943            & 0.9955 \\
\cmidrule{2-7}
&\multirow{4}{*}{LS} 
    & Avg.        & 0.7837          & \textbf{0.7745}            & 0.7790            & 0.7971 \\  
    && St. Dev.    & 0.3759          & 0.3851            & 0.3818            & \textbf{0.3495} \\
    && Min.       & 0.0294          & \textbf{0.0057}   & 0.0124            & 0.0373 \\
    && Max.       & \textbf{1.0000} & \textbf{1.0000}   & \textbf{1.0000}   & \textbf{1.0000} \\
\midrule
\multirow{7}{*}{TSR} 
&\multirow{4}{*}{MS} 
    & Avg.       & 0.4770           & 0.5529            & 0.4592           & \textbf{0.5563} \\  
    && St. Dev.   & 0.4274           & 0.3757            & 0.4402           & \textbf{0.3745} \\
    && Min.      & 0.0520           & 0.1154            & 0.0419           & \textbf{0.1324} \\
    && Max.      & \textbf{1.0000}  & \textbf{1.0000}   & \textbf{1.0000}  & \textbf{1.0000} \\
\cmidrule{2-7}
&\multirow{4}{*}{LS} 
    & Avg.       & 0.4066           & \textbf{0.3919}            & 0.3989           & 0.4377 \\  
    && St. Dev.   & 0.4825           & 0.4816            & 0.4799           & \textbf{0.4704} \\
    && Min.      & 0.0023           & \textbf{0.0000}   & 0.0023           & 0.0181 \\
    && Max.      & \textbf{1.0000}  & \textbf{1.0000}   & \textbf{1.0000}  & \textbf{1.0000} \\
\bottomrule
\end{tabular}
\end{table}

\tablename{}~\ref{tab:all-you-need-is-love} provides more detailed MS and LS testing results for the chosen similarity sources. This table also presents the results obtained for the WordNet-based similarity measure (WUP), which is our reference. Labels chosen with MS strategies based on all similarity sources are significantly easier to reach than those picked with LS strategies for all target vision models. This is evident in weaker adversarial attacks, such as CW or FGM. For example, the TSR value for EfficientNetV2B0 was 0.3 and 0.04 for MS and LS, respectively. For mighty attacks, PGD in our case, the impact of the label choice is negligible, i.e., the attack success rate oscillates around 100\%. These results suggest that the language and image-language models essentially share similar perceptions between classes with pure vision models. Thus, they are good candidates for target selection and further effectiveness evaluation.

While TSR consistently reflects the effectiveness of target label selection, a closer inspection of the FR distributions reveals that label semantics also subtly affect non-targeted success. This is especially evident in the violin's minimum values and lower tails in \figurename{}~\ref{fig:bar-plots_MSLS}, where MS variants exhibit visibly truncated lower ranges than LS variants. Although the effect on FR is not as pronounced as on TSR, it remains measurable and indicates that semantic proximity plays a role even in less strict attack goals. Such differences, particularly in the bottom quantiles of FR, suggest that label selection influences overall attack difficulty beyond target-hitting accuracy alone.

\begin{table}[!ht]
\caption{%
Attack performance metrics for target label selection using BERT embeddings, TinyLLAMA-based target label selection, CLIP-based semantic similarity, and WordNet-based WUP similarity.
Fooling Rate (FR) and Targeted Success Rate (TSR) are reported for both Most Similar (MS) and Least Similar (LS) cases across five attack methods (CW, FGM, MIM, PGD, and SPSA) and three target models (EfficientNetV2-B0, MobileNetV2, and ResNet50V2). The best values for each attack-variant pair are highlighted in bold.
}\label{tab:all-you-need-is-love}
\centering 
\begin{tabular}{@{}*{2}{c@{ }}c*{6}{c}@{}}
    \toprule
    \multirow{2}{*}{\bf Model} &
    \multirow{2}{*}{\bf Attack} & \multirow{2}{*}{\bf Var.} & \multicolumn{2}{@{ }c@{ }}{\bf EfficientNetV2-B} & \multicolumn{2}{@{ }c@{ }}{\bf MobileNetV2} & \multicolumn{2}{@{ }c@{ }}{\bf ResNet50V2} \\
    &&  & \textbf{FR} & \textbf{TSR} & \textbf{FR} & \textbf{TSR} & \textbf{FR} & \textbf{TSR} \\
    \midrule
\multirow{12}{*}{BERT} &
    \multirow{2}{*}{CW} & MS & 0.15 & 0.15 & 0.32 & 0.32 & 0.12 & 0.12 \\
    && LS & {0.03} & {0.03} & 0.13 & {0.13} & 0.03 & 0.03 \\
    \cmidrule{2-9}
    &\multirow{2}{*}{FGM} & MS & 0.89 & 0.13 & \textbf{0.93} & 0.09 & {0.91} & 0.05 \\
    && LS & \textbf{0.88} & \textbf{0.01} & \textbf{0.94} & \textbf{0.00} & \textbf{0.92} & \textbf{0.00} \\
    \cmidrule{2-9}
    &\multirow{2}{*}{MIM} & MS & 0.99 & 0.99 & 0.99 & \textbf{1.00} & 0.97 & 0.88 \\
    && LS & 1.00 & 1.00 & 1.00 & {1.00} & 0.98 & {0.86} \\
    \cmidrule{2-9}
    &\multirow{2}{*}{PGD} & MS & \textbf{1.00} & \textbf{1.00} & 0.99 & \textbf{1.00} & {0.99} & {0.99} \\
    && LS & \textbf{1.00} & \textbf{1.00} & \textbf{1.00} & \textbf{1.00} & \textbf{1.00} & \textbf{0.99} \\
    \cmidrule{2-9}
    &\multirow{2}{*}{SPSA} & MS & 0.91 & 0.10 & \textbf{0.97} & 0.21 & 0.94 & 0.11 \\
    && LS & 0.93 & 0.01 & 0.98 & {0.03} & 0.95 & \textbf{0.00} \\
\midrule
\multirow{12}{*}{CLIP} &
    \multirow{2}{*}{CW} & MS & 0.27 & 0.28 & \textbf{0.47} & \textbf{0.47} & \textbf{0.26} & \textbf{0.26} \\
    && LS & \textbf{0.01} & \textbf{0.01} & \textbf{0.09} & \textbf{0.09} & \textbf{0.01} & \textbf{0.01} \\
    \cmidrule{2-9}
    &\multirow{2}{*}{FGM} & MS & 0.88 & \textbf{0.27} & {0.92} & \textbf{0.19} & 0.90 & 0.12 \\
    && LS & \textbf{0.88} & \textbf{0.01} & \textbf{0.94} & \textbf{0.00} & \textbf{0.92} & \textbf{0.00} \\
    \cmidrule{2-9}
    &\multirow{2}{*}{MIM} & MS & 0.99 & 0.99 & 0.99 & \textbf{1.00} & \textbf{0.98} & 0.94 \\
    && LS & {1.00} & {0.97} & {1.00} & {1.00} & {0.98} & {0.80} \\
    \cmidrule{2-9}
    &\multirow{2}{*}{PGD} & MS & 0.99 & \textbf{1.00} & 0.99 & \textbf{1.00} & \textbf{0.99} & \textbf{0.99} \\
    && LS & \textbf{1.00} & \textbf{1.00} & \textbf{1.00} & \textbf{1.00} & \textbf{1.00} & \textbf{0.99} \\
    \cmidrule{2-9}
    &\multirow{2}{*}{SPSA} & MS & 0.90 & 0.20 & \textbf{0.98} & 0.36 & 0.94 & \textbf{0.23} \\
    && LS & \textbf{0.90} & \textbf{0.00} & \textbf{0.97} & \textbf{0.01} & \textbf{0.93} & \textbf{0.00} \\
\midrule
\multirow{12}{*}{LLAMA} &
    \multirow{2}{*}{CW} & MS & 0.10 & 0.11 & 0.28 & {0.29} & 0.09 & 0.10 \\
    && LS & \textbf{0.01} & \textbf{0.01} & 0.11 & {0.11} & \textbf{0.01} & \textbf{0.01} \\
    \cmidrule{2-9}
    &\multirow{2}{*}{FGM} & MS & 0.87 & 0.10 & \textbf{0.93} & 0.07 & {0.91} & 0.04 \\
    && LS & {0.89} & \textbf{0.01} & 0.95 & \textbf{0.00} & \textbf{0.92} & \textbf{0.00} \\
    \cmidrule{2-9}
    &\multirow{2}{*}{MIM} & MS & 0.99 & 0.99 & 0.99 & \textbf{1.00} & {0.98} & 0.87 \\
    && LS & {0.99} & 0.97 & 1.00 & {1.00} & {0.98} & {0.82} \\
    \cmidrule{2-9}
    &\multirow{2}{*}{PGD} & MS & 0.99 & \textbf{1.00} & 0.99 & \textbf{1.00} & {0.99} & {0.99} \\
    && LS & \textbf{1.00} & \textbf{1.00} & \textbf{1.00} & \textbf{1.00} & \textbf{1.00} & \textbf{0.99} \\
    \cmidrule{2-9}
    &\multirow{2}{*}{SPSA} & MS & 0.91 & 0.07 & \textbf{0.97} & 0.16 & 0.94 & 0.09 \\
    && LS & {0.91} & {0.01} & \textbf{0.97} & {0.03} & {0.95} & {0.01} \\
\midrule
\multirow{12}{*}{WUP} &
    \multirow{2}{*}{CW} & MS & \textbf{0.29} & \textbf{0.30} & 0.46 & 0.46 & 0.23 & 0.24 \\
    && LS & {0.04} & {0.04} & 0.24 & 0.24 & 0.11 & 0.11 \\
    \cmidrule{2-9}
    &\multirow{2}{*}{FGM} & MS & \textbf{0.91} & \textbf{0.27} & \textbf{0.93} & 0.18 & \textbf{0.91} & \textbf{0.13} \\
    && LS & \textbf{0.88} & {0.02} & \textbf{0.94} & 0.04 & \textbf{0.92} & {0.02} \\
    \cmidrule{2-9}
    &\multirow{2}{*}{MIM} & MS & \textbf{1.00} & \textbf{1.00} & \textbf{1.00} & \textbf{1.00} & \textbf{0.98} & \textbf{0.95} \\
    && LS & 1.00 & 1.00 & 1.00 & 1.00 & 0.99 & {0.95} \\
    \cmidrule{2-9}
    &\multirow{2}{*}{PGD} & MS & \textbf{1.00} & \textbf{1.00} & \textbf{1.00} & \textbf{1.00} & \textbf{0.99} & \textbf{0.99} \\
    && LS & \textbf{1.00} & \textbf{1.00} & \textbf{1.00} & \textbf{1.00} & \textbf{1.00} & \textbf{0.99} \\
    \cmidrule{2-9}
    &\multirow{2}{*}{SPSA} & MS & \textbf{0.92} & \textbf{0.22} & \textbf{0.98} & {0.38} & \textbf{0.95} & \textbf{0.23} \\
    && LS & 0.92 & \textbf{0.02} & 0.98 & 0.09 & 0.94 & 0.04 \\
\bottomrule
\end{tabular}
\end{table}

\subsection{Comparing Similarity Sources for Target Label Selection}

We further analyze the results in \figurename{}~\ref{fig:bar-plots_MSLS} through \tablename{}~\ref{tab:all-you-need-is-love} to determine the use cases where particular similarity sources are better for testing. 
Text and text-image models produced harder-to-reach LS targets than WUP, as evidenced by lower FR and TSR values. These results support the view that embedding-based similarities better capture global semantic distance.
Thus, it aligns with our observation and motivation: WordNet-based similarity measures can lack enough variability when they are more distant from the ground truth due to many similar values outside the closest neighborhood. Besides, in the case of the MS variant, WordNet-based WUP similarity achieved the best results (Avg. TSR of 0.56), closely followed by the CLIP text-image model (Avg. TSR of 0.55). They also reached the same Avg. FR value. Moreover, BERT and TinyLLAMA also obtained high FR values (i.e., 0.8 and 0.81), but the TSR results exhibit that their determined most similar labels are slightly more challenging to reach. 

Considering the above, the overall results suggest the following practical implications: Text-image models and WordNet-based measures can be slightly better for local similarity estimation. Therefore, to construct best-case scenarios for security testing, using text-image models as an alternative to WordNet semantics is best. Furthermore, text and text-language models are superior at estimating global similarity (i.e., constructing worst-case scenarios).

\subsection{Impact of Similarity Sources on Attack Severity}

\begin{figure}[!htp]
  \centering
\subfloat[Most Similar (MS)]{%
    \includegraphics[%
    trim={10pt 10 10pt 10pt}, clip, 
    width=\wpr\linewidth]{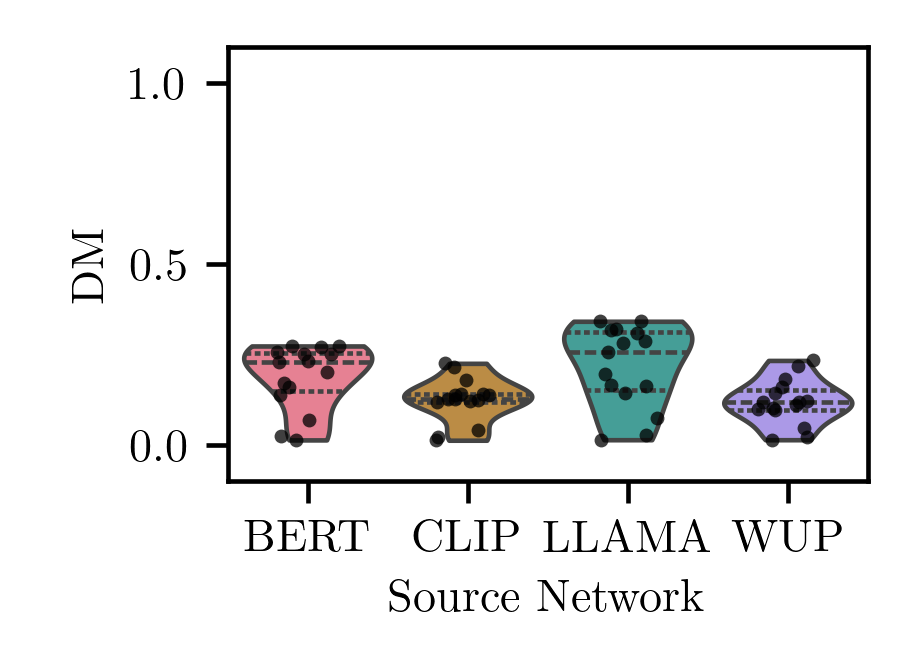}
  }%
  \subfloat[Least Similar (LS)]{%
    \includegraphics[%
    trim={10pt 10 10pt 10pt}, clip, 
    width=\wpr\linewidth]{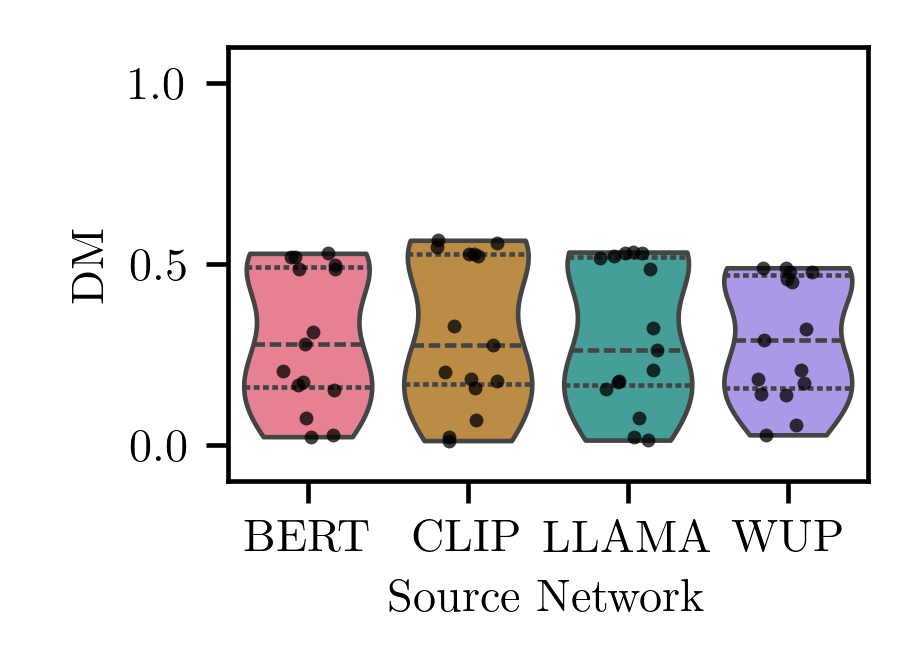}
  }
  \caption{%
  Average Dissimilarity Metric (DM) values for all semantic similarity sources under the (a) MS and (b) LS selection strategies. DM measures the semantic distance between post-attack predictions and ground truth labels in the model's class space. Lower values in (a) and higher values in (b) are desirable and indicate well-aligned similarity assessments. Dashed lines denote the quartile ranges, and black dots represent individual samples.
  }
  \label{fig:bar_plots_DM}
\end{figure}

\figurename{}~\ref{fig:bar_plots_DM} presents the DM values obtained for a given similarity source in both MS and LS test case scenarios. These values allow us to assess the overall impact of attacks performed with a given target on the model prediction, regardless of the targeted attack's success. Even in the absence of a successful targeted hit, the DM quantifies how far the prediction drifts in the semantic class space. For the MS variant, the average DM value should be low, i.e., labels are perceived as close to the ground truth. Conversely, for the LS variant, high values imply that labels are perceived as distant from the reference. Considering this, the DM results corroborate our finding about the superiority of text and text-image models in selecting the LS labels. Specifically, the best average result obtained for CLIP is approximately 0.31 compared to 0.30 for purely text models and 0.29 for WUP. St. Dev. values are comparable for all source networks. In the case of the MS variant, CLIP again obtained similar results to WUP, whereas the purely text models obtained inferior results with higher dispersion. This shows that WordNet measures and text-image models are more suitable for local similarity estimation. Text and vision-language models are better suited for identifying semantically distant targets, confirming their utility in constructing robust worst-case scenarios.

To support these inferences, \tablename{}~\ref{tab:dm_std_min_max} displays the St. Dev., Min., and Max. values of the DM values. These values are much more variable for the LS testing scenarios. Such a fact is reasonable, as the desired high values in this strategy are much more challenging to reach than those in the MS scenario. The variability, again, is lower for the WUP and CLIP (text-image model) used as similarity sources. The minimum values show that even in the LS scenarios, the predicted labels can still be relatively close to the ground truth for some images and weaker attacks. However, the maximum DM values obtained for the LS variant are about 0.5 (0.6 for CLIP), and approximately 0.2/0.3 for the MS variant. This means the LS variant can result in much more distant predictions than the MS one, by about 1/3 of the total rank distance in the model's class space. 

\begin{table}[ht]
\centering
\caption{Aggregated statistics for the Dissimilarity Metric (DM) for both Most Similar (MS) and Least Similar (LS) variants. For MS cases, higher values imply easier-to-reach adversarial targets, while for LS cases, lower values indicate more challenging targets. The best results per row are highlighted in bold across the sources.}%
\label{tab:dm_std_min_max}
\begin{tabular}{llcccc}
\toprule
\textbf{Variant} & \textbf{Statistic} & \textbf{BERT} & \textbf{CLIP} & \textbf{LLAMA}  & \textbf{WUP} \\
\midrule
\multirow{4}{*}{MS} 
    & Avg.       & 0.1881  & 0.1253           & \textbf{0.2164}  & 0.1197 \\
    & St. Dev.   & 0.0898  & \textbf{0.0613}  & 0.1131          & 0.0632            \\
    & Min.       & 0.0143  & 0.0134           & \textbf{0.0148} & \textbf{0.0148}   \\
    & Max.       & 0.2739  & 0.2259           & \textbf{0.3422} & 0.2339            \\
\midrule
\multirow{4}{*}{LS} 
    & Avg.       & 0.2969  & 0.3119          & 0.3027           & \textbf{0.2924} \\
    & St. Dev.   & 0.1933  & 0.2111          & 0.2008           & \textbf{0.1710}   \\
    & Min.       & 0.0233  & \textbf{0.0124} & 0.0138           & 0.0283            \\
    & Max-       & 0.5300  & 0.5661          & 0.5334           & \textbf{0.4901}   \\
\bottomrule
\end{tabular}
\end{table}

\subsection{Compatibility of target labels with network perception}

\begin{figure}[!htp]
  \centering
  \subfloat[Most Similar (MS)]{%
    \includegraphics[%
    trim={10pt 0 10pt 10pt}, clip,
    width=\wpr\linewidth]{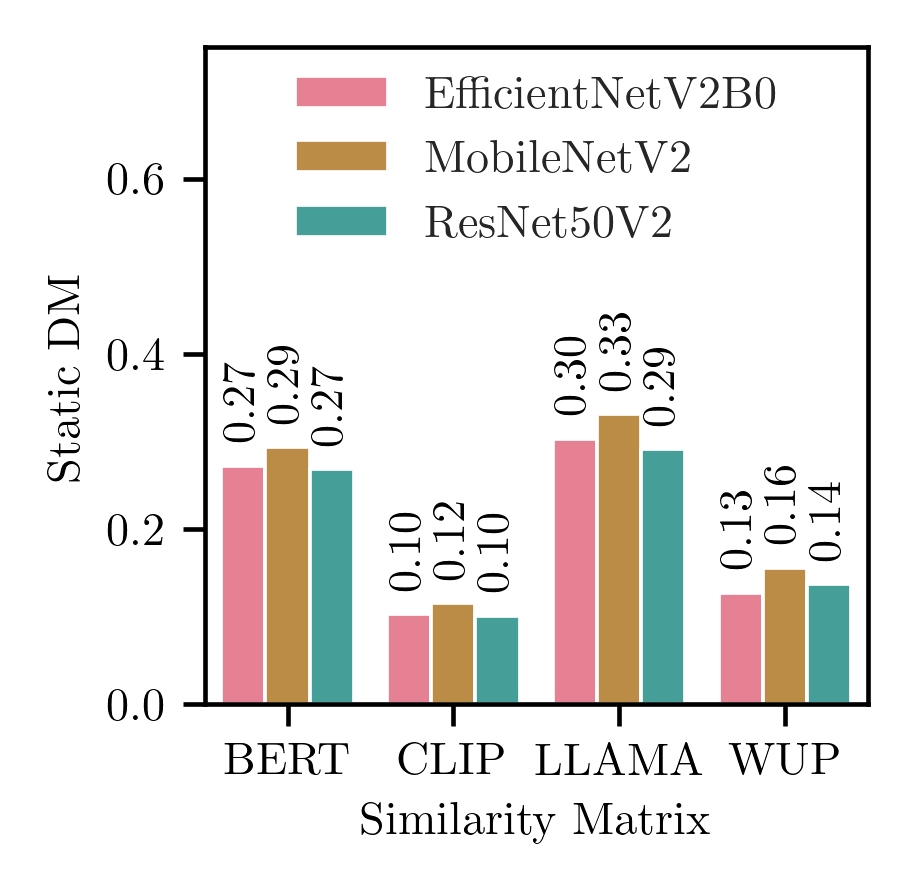}
  }%
  \subfloat[Least Similar (LS)]{%
    \includegraphics[%
    trim={10pt 0 10pt 10pt}, clip,
    width=\wpr\linewidth]{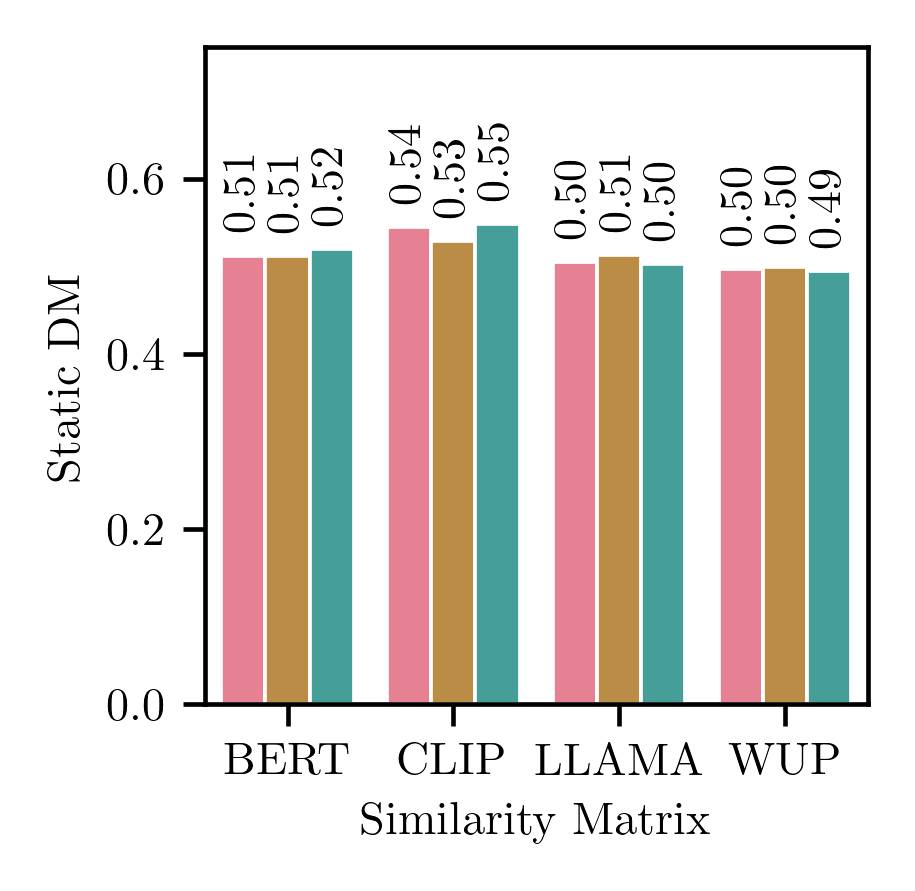}
  }
  \caption{Static Dissimilarity Metric (DM) values computed for all ImageNet classes, based on Most Similar (MS) and Least Similar (LS) target labels obtained from different semantic similarity sources. In ImageNet's 1000-class space context, the optimal DM values are one thousandth for the MS variant and one for the LS variant, reflecting complete alignment with the model's perceived inter-class similarity. 
  }
  \label{fig:bar-plots_MSLS_static}
\end{figure}

\figurename{}~\ref{fig:bar-plots_MSLS_static} presents the static DM scores computed between ground truth labels and target labels selected via different semantic similarity sources, i.e., without using any image data. These scores reflect the mean normalized rank distance from the ground truth in the model's output space. We observe consistent patterns by comparing them with the post-attack DM results in \figurename{}~\ref{fig:bar_plots_DM}. For example, in static and predictive settings, TinyLLAMA produces the highest DM in the MS variant. It indicates that its most similar targets are semantically less aligned than those from other sources. Conversely, CLIP and WUP consistently yield the lowest DM values in the MS case, reflecting better local similarity alignment. These trends are also reflected in FR and TSR scores from \tablename{}~\ref{tab:all-you-need-is-love}, where CLIP achieves top performance for MobileNetV2 and ResNet50V2. CLIP again produces the highest static DM values for the LS variant, confirming its effectiveness in identifying worst-case adversarial targets.

These results suggest that static DM can assess, \textit{a priori}, the alignment between semantic similarity sources and a vision model's internal class structure. The close match between static and predictive DM trends confirms that this metric provides a reliable compatibility measure for evaluating and selecting semantic sources.

\def\vari{_new}

\section{Conclusions}

In this work, we proposed a unified framework for semantics-guided adversarial target label selection that leverages language and vision language models to estimate semantic similarity between class labels. This approach enables constructing interpretable and systematic best and worst-case adversarial scenarios. Unlike model or image-dependent methods, our strategy provides a reproducible and scalable alternative that decouples target selection from confounding instance-level factors.

Our experimental evaluation, covering three similarity source models, three vision architectures, and five attack methods, demonstrated that pretrained text and text image models are highly effective in guiding adversarial target selection. The text image model CLIP consistently outperformed the WordNet-based baseline in identifying semantically distant, hard-to-reach labels and performed on par with it in best-case scenarios. Furthermore, the Dissimilarity Metric applied to post-attack predictions revealed that text and vision language embeddings offer a more faithful estimation of global class relationships. At the same time, CLIP and WordNet remain competitive in capturing local similarities.

A noteworthy insight from our analysis is that the static Dissimilarity Metric, computed without requiring image inputs, can serve as an early indicator of how compatible a given similarity source is with a vision model's internal class perception. The strong alignment between static and predictive metric trends suggests that model vulnerability can, to some extent, be assessed in advance, enabling proactive security diagnostics before testing begins.

Our findings support using semantics-aware target selection via pretrained models as a practical, flexible, and effective alternative to traditional static lexical databases. Our method improves transparency and standardization in adversarial evaluation, laying the groundwork for more reproducible testing pipelines and interpretable security benchmarks. Looking forward, we aim to extend this framework to adjust contextual and image-conditioned semantics and adapt it to multi-label and open vocabulary classification, further broadening its applicability in real-world settings.

\section*{Acknowledgments}

This work was supported by the~START Scholarship of the Foundation for Polish Science (FNP) for outstanding young scholars, agreement No START 017.2025.  

\noindent
The authors utilised OpenAI's ChatGPT‑4o and Grammarly for language polishing under human supervision. No AI was used for scientific ideas, data analysis, or interpretation. The authors remain fully responsible for the manuscript.

\bibliography{IEEEabrv, refs}
\bibliographystyle{IEEEtran}

\end{document}